%% file: main.tex
\documentclass[10pt]{article} 
\usepackage[accepted]{tmlr}

\input{math_commands.tex}

\usepackage{hyperref}
\usepackage{url}
\usepackage{graphicx}
\usepackage{booktabs}
\usepackage{caption}

\title{Causal-Discovery Performance of ChatGPT in the context of Neuropathic Pain Diagnosis}

\author{\name Ruibo Tu \email ruibo@kth.se \\
      KTH Royal Institute of Technology
      \AND
      \name Chao Ma \email chaoma@microsoft.com \\
      \addr Microsoft Research 
    \AND
    \name Cheng Zhang \email cheng.zhang@microsoft.com\\
      Microsoft Research
      }

\begin{document}

\maketitle

\paragraph{Introduction.}
ChatGPT\cite{chatgpt} has demonstrated exceptional proficiency in natural language conversation, e.g., it can answer a wide range of questions while no previous large language models can. Thus, we would like to push its limit and explore its ability to answer causal discovery questions by using a medical benchmark \cite{tu2019neuropathic} in causal discovery. 

Causal discovery aims to uncover the underlying unknown causal relationships based purely on observational data\cite{glymour2019review}. 
In contrast, applying ChatGPT to answer the questions about causal relationships is fundamentally different.
With the current version of ChatGPT, we can only use the names (meta information) instead of observational data  of variables to answer causal questions. The answers to the causal questions given by ChatGPT are based on a trained large language model, which can be viewed as an approximation for existing knowledge in the training natural language data. Nevertheless, such investigations still provide us valuable insights into ChatGPT and raise more thoughts about how to leverage its ability.  
But we need to exercise great caution in the conclusion as benchmarks \cite{pair, tu2019neuropathic} utilizing known knowledge are set for evaluation purposes instead of the goal of the causal discovery.

\begin{minipage}{0.6\textwidth}
    \centering
    \includegraphics[width=0.7\textwidth]{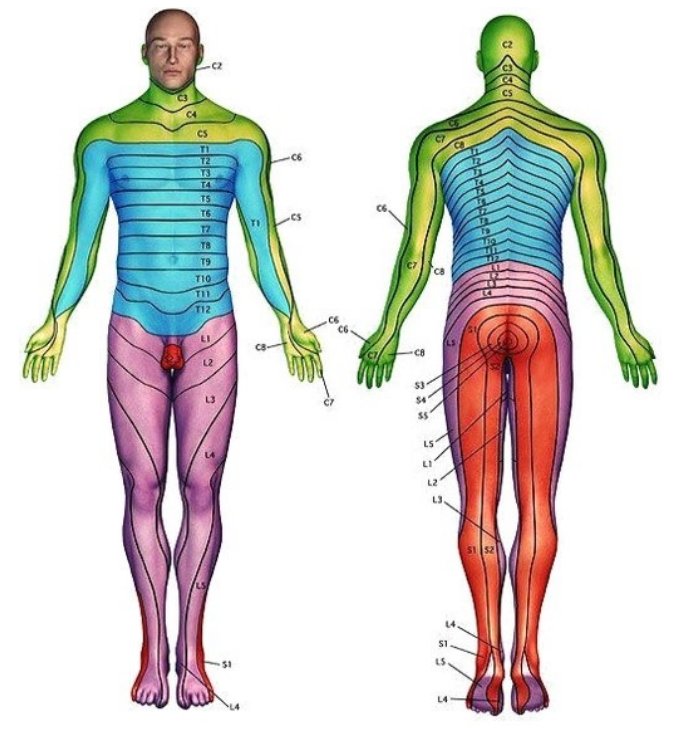}
\captionof{figure}{Dermatome map \cite{map} as a reference for this benchmark.}
    \label{fig:dermatome_map}
\end{minipage}
\begin{minipage}{0.34\textwidth}
    \centering
    \begin{tabular}{c c c}
    \toprule
Precision &	Recall &	F-score\\
\midrule
1&	0.12	& 0.2142857143\\
\bottomrule
    \end{tabular}
\captionof{table}{Test results demonstrate high precision and low recall.}
    \label{tab:fscore}
    \vspace{1cm}
    \centering
    \begin{tabular}{c c c}
    \toprule
	& Negative &	Positive\\
Negative	&50&	44\\
Positive	&0&	6\\
\bottomrule
    \end{tabular}
 \captionof{table}{Confusion matrix showing that there were no false positives. Rows are predictions and columns are ground truth.}
    \label{tab:confusionmatrix}
\end{minipage}

\begin{figure}
    \centering
    \includegraphics[width=0.9\textwidth]{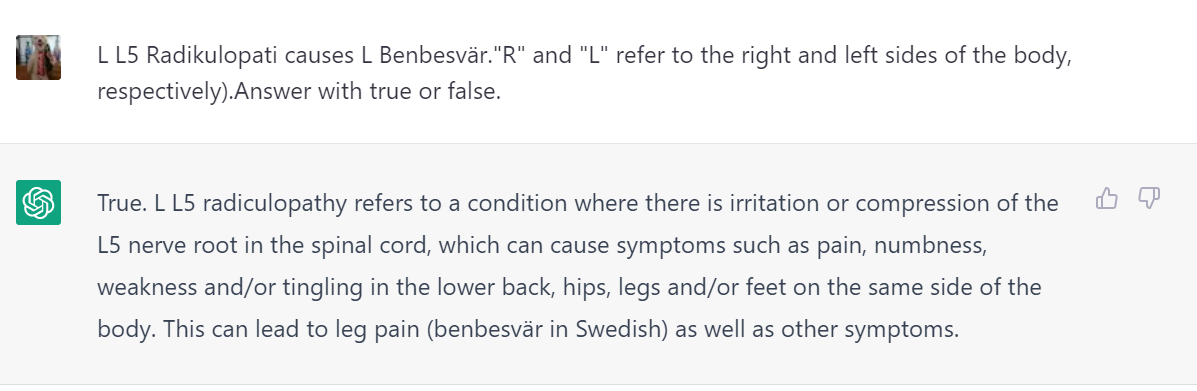}
    \caption{Example showing that ChatGPT can correctly answer the question and provide reasonable explanations.}
    \label{fig:swedish}
\end{figure}

\begin{figure}
    \centering
    \includegraphics[width=0.9\textwidth]{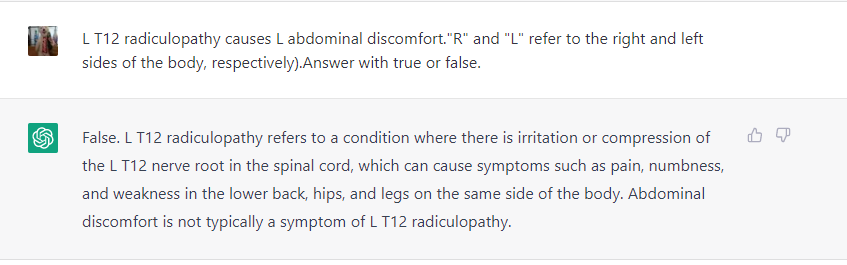}
    \caption{ The lower abdominal is the region where T12 nerve passes. If looking at the dermatome map \ref{fig:dermatome_map}, 
    it is easy to identify lower back, hip, and lower abdominal discomfort can all be caused by T12 radiculopathy.}
    \label{fig:lack_knowledge}
\end{figure}

\begin{figure}
    \centering
    \includegraphics[width=0.9\textwidth]{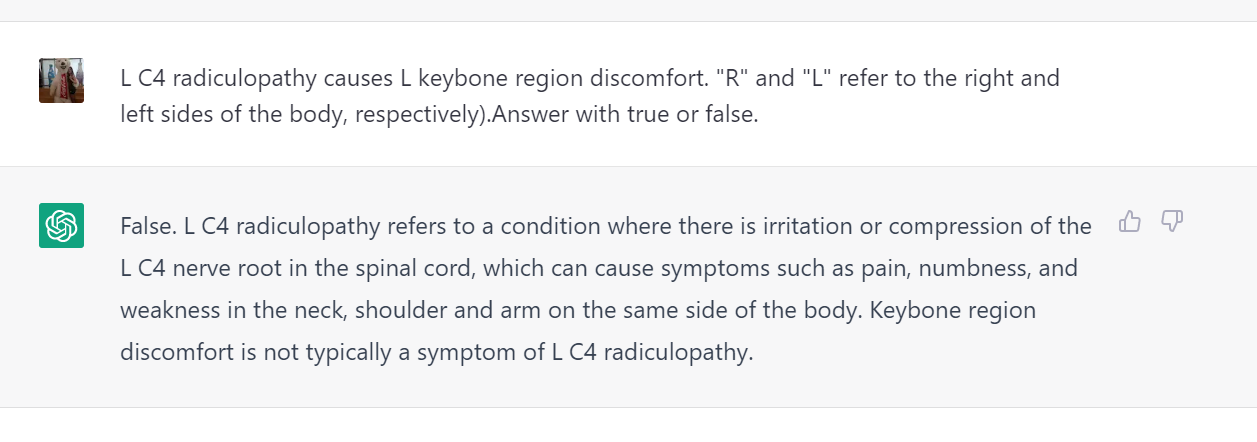}
    \includegraphics[width=0.9\textwidth]{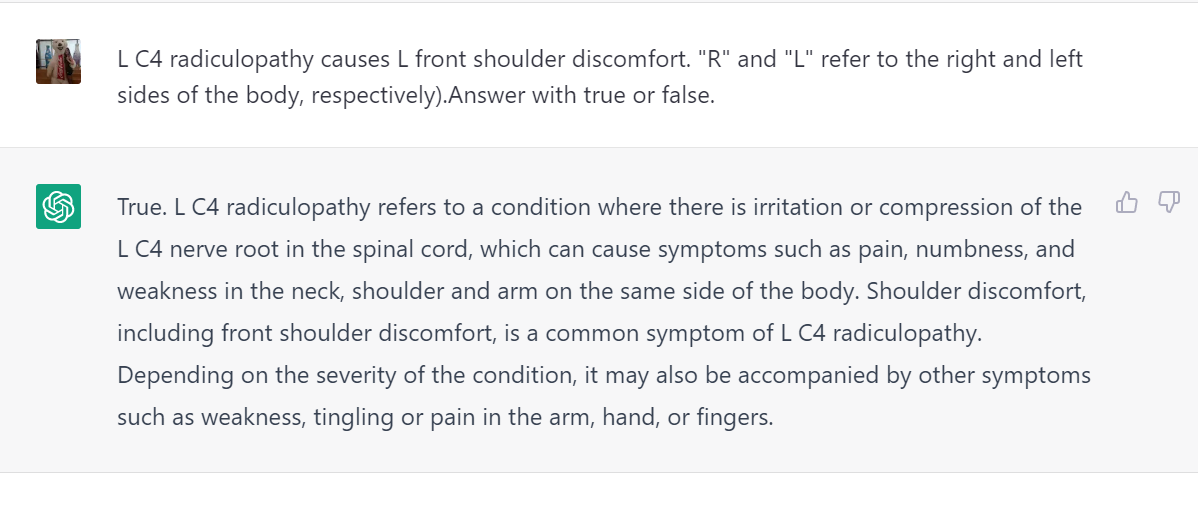}
    \caption{Example showing that ChatGPT fails to understand the region on the body. The area around the key bone is largely overlapping with the front shoulder area especially when the patient describes the symptoms. }
    \label{fig:region_understanding}
\end{figure}

\paragraph{Results and Insights.}
The ground-truth causal relationships in neuropathic pain diagnosis are obtained from both a domain expert and known medical literature \cite{tu2019neuropathic}. 
As the number of all possible cause-effect pairs in this context is huge (more than $10000$ pairs), we cannot test all of them manually. Thus, we sub-sampled 50 positive pairs (ground-true causal relationships) and 50 negative pairs (wrong causal relationships) from the dataset and generated the question in the format of "X causes Y. 
Answer true or false", where X and Y are sampled pairs from the full causal map of the neuropathic pain dataset. The full test results can be found at \url{shorturl.at/amBX1}. 
Many individual answers are reasonable, such as in Figure \ref{fig:swedish}, but the performance is still flowed currently. As shown in Table \ref{tab:fscore} and \ref{tab:confusionmatrix}, ChatGPT tends to make false negative mistakes. We inspected the results qualitatively and quantitatively and observed that:

It only understands the languages that are typically used to describe the situations but not the underlying knowledge. We provide two examples to demonstrate it. The first example is shown in Figure \ref{fig:lack_knowledge}. It cannot identify the lower abdominal discomfort that can be caused by T12 radiculopathy. The explanation identifies the lower back, hip, and leg region only, while T12 nerve goes through these regions shown in Figure \ref{fig:dermatome_map}, and the lower abdominal region is part of it. Thus, it indicates that it provides the answer based on the trained content but does not understand the human body's nervous system. The second example is shown in Figure \ref{fig:region_understanding}, which demonstrates a lack of understanding of how regional discomfort is described. The region around the key bone is the upper shoulder region. ChatGPT can identify shoulder discomfort as an effect but not the discomfort around the key bone. 

\begin{figure}
    \centering
    \includegraphics[width=0.9\textwidth]{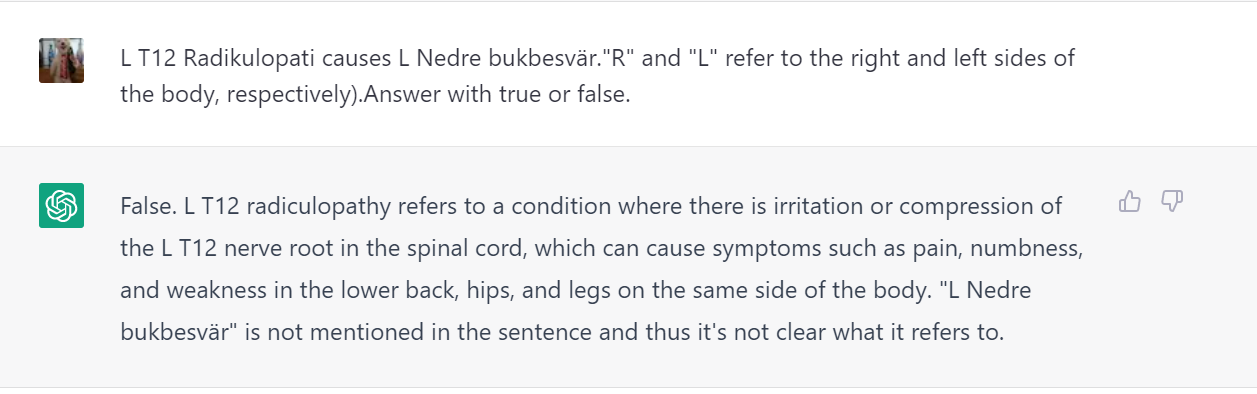}
    \caption{Example showing that ChatGPT can identify foreign language time by time and that it is not very reliable. }
    \label{fig:noswedish}
\end{figure}

Its performance is not yet consistent and not stable. Firstly, we observe that it provides different answers to the same question. We have tested some of the queries twice on different days. As shown in Table \ref{tab:twoday}, the answers on the first day differ from the ones on the second day significantly. The answers on the second day are much more conservative to claim a causal relationship. This may be due to internal model updates. Such inconsistent performance is a major concern for answering causal questions. As the later results have very few positive answers, the final results that we used considered the earlier results when available for the table \ref{tab:fscore} and \ref{tab:confusionmatrix}. Secondly, as the original dataset is associated with terms in Swedish, we found that ChatGPT can correctly identify Swedish in some cases, such as in Figure \ref{fig:swedish}, but fails in some other cases, such as in Figure \ref{fig:noswedish}. This may contribute further to a large number of false negatives.

\paragraph{Conclusion.} Based on the observations, we find:

\begin{itemize}
\item There are some limitations for the current ChatGPT in terms of understanding new concepts and knowledge beyond the existing corpus of text training data. Moreover, the consistency and stability of its performance need to be improved. Such improvements can happen without a paradigm shift in the models. 
\item We need to be extremely cautious about using causal claims made by ChatGPT as causal discovery results. This is because causal discovery and causal question answering with large language models are fundamentally different tasks. Causal benchmarks may be biased towards utilizing existing knowledge for evaluation \cite{tu2019neuropathic, pair}, which is against the goal of causal discovery.   
\item In some situations, ChatGPT does give correct answers that can be non-trivial to obtain from a domain expert, which could serve as a good complementary for causal discovery methods to resolve corner cases. This might open up new research opportunities for the causal community on utilizing the recent developments of large language models to complement, improve and develop better causal machine learning tools. 
\end{itemize}

Although there are existing limitations, we believe that opportunities for ChatGPT can help to improve the causality research is huge. With deep integration with ChatGPT type of models and interface. We can also imagine a future where ChatGPT can answer different causal questions.

\clearpage

\begin{table}[]
    \centering
    \begin{tabular}{p{11cm} p{1.8cm} p{0.7cm} p{0.7cm}}
    \toprule
Query &	GroundTruth from Expert & Jan18 Test	& Jan19 Test \\
\midrule
L L4 Radikulopati causes R Höftkamsbesvär."R" and "L" refer to the right and left sides of the body, respectively).Answer with true or false.&	TRUE&	0	&0 \\
L T12 Radikulopati causes L Nedre bukbesvär."R" and "L" refer to the right and left sides of the body, respectively).Answer with true or false.	&TRUE	&0	&0\\
R L1 Radikulopati causes L Ljumskbesvär."R" and "L" refer to the right and left sides of the body, respectively).Answer with true or false.	&TRUE	&0&	0\\
R C6 Radikulopati causes L Laterala armsbesvär."R" and "L" refer to the right and left sides of the body, respectively).Answer with true or false.	&TRUE	&1&	0\\
R C6 Radikulopati causes R Armbågsbesvär."R" and "L" refer to the right and left sides of the body, respectively).Answer with true or false.	&TRUE	&1&	0\\
R L1 Radikulopati causes IBS."R" and "L" refer to the right and left sides of the body, respectively).Answer with true or false.	&TRUE &	0&	0 \\
L C5 Radikulopati causes Nackbesvär."R" and "L" refer to the right and left sides of the body, respectively).Answer with true or false.	&TRUE	&1&	0 \\
R C5 Radikulopati causes R Laterala armbågsbesvär."R" and "L" refer to the right and left sides of the body, respectively).Answer with true or false.	&TRUE	&0	&0 \\
L C6 Radikulopati causes R Handbesvär."R" and "L" refer to the right and left sides of the body, respectively).Answer with true or false.	&TRUE	&1&	0 \\
L L4 Radikulopati causes L Laterala vadbesvär."R" and "L" refer to the right and left sides of the body, respectively).Answer with true or false.	&TRUE	&0&	0 \\
R S1 Radikulopati causes R Lårbesvär."R" and "L" refer to the right and left sides of the body, respectively).Answer with true or false.	&TRUE	&1&	0 \\
L T5 Radikulopati causes L Bröstbesvär."R" and "L" refer to the right and left sides of the body, respectively).Answer with true or false.	&TRUE	&0	&0 \\
L C5 Radikulopati causes R Interskapulära besvär."R" and "L" refer to the right and left sides of the body, respectively).Answer with true or false.	&TRUE	&0	&0 \\
R C6 Radikulopati causes R Under armsbesvär."R" and "L" refer to the right and left sides of the body, respectively).Answer with true or false.	&TRUE	&0	&0 \\
L L1 Radikulopati causes L Mediala ljumskbesvär."R" and "L" refer to the right and left sides of the body, respectively).Answer with true or false.	&TRUE	&0&	0 \\
L L1 Radikulopati causes L Adduktortendalgi."R" and "L" refer to the right and left sides of the body, respectively).Answer with true or false.	&TRUE	&0&	0 \\
L T10 Radikulopati causes IBS."R" and "L" refer to the right and left sides of the body, respectively).Answer with true or false.&	TRUE&	0	&0\\
R L5 Radikulopati causes L Bakhuvudvärk."R" and "L" refer to the right and left sides of the body, respectively).Answer with true or false.&	TRUE	&0&	0 \\
\bottomrule
    \end{tabular}
    \caption{Results demonstrate lack of consistency using ChatGPT.}
    \label{tab:twoday}
\end{table}

\clearpage
\bibliography{tmlr}
\bibliographystyle{abbrv}


\end{document}

%% file: math_commands.tex
\usepackage{amsmath,amsfonts,bm}

\def\eqref#1{equation~\ref{#1}}

\def\1{\bm{1}}

\DeclareMathAlphabet{\mathsfit}{\encodingdefault}{\sfdefault}{m}{sl}
\SetMathAlphabet{\mathsfit}{bold}{\encodingdefault}{\sfdefault}{bx}{n}

